\documentclass[letterpaper, 10 pt, conference]{ieeeconf}  

\IEEEoverridecommandlockouts                              

\usepackage[pdftex]{graphicx}
\usepackage{algorithm}
\usepackage{algpseudocode}

\usepackage{amssymb,amsmath, amsthm}
\usepackage{subfig}
\usepackage{indentfirst}
\usepackage{color}
\usepackage[hidelinks]{hyperref}
\usepackage{enumerate}
\usepackage{tikz}
\usetikzlibrary{shapes,arrows}
\usepackage{epstopdf}
\usepackage{epsfig}
\usepackage{wrapfig}
\usepackage[font=small]{caption}
\usepackage{balance}
\usepackage{setspace}

\def\mf{\mathbf}
\def\mb{\mathbb}
\def\mc{\mathcal}
\def\beq{\begin{equation*}}
\def\eeq{\end{equation*}}
\def\bql{\begin{equation}}
\def\eql{\end{equation}}
\def\bqn{\begin{eqnarray*}}
\def\eqn{\end{eqnarray*}}
\def\bnl{\begin{eqnarray}}
\def\enl{\end{eqnarray}}
\def\bna{\bql\begin{array}{rcl}}
\def\ena{\end{array}\eql}
\def\bnn{\beq\begin{array}{rcl}}
\def\enn{\end{array}\eeq}
\def\bma{\begin{bmatrix}}
\def\ema{\end{bmatrix}}
\def\bmx{\begin{matrix}}
\def\emx{\end{matrix}}
\def\ben{\begin{enumerate}}
\def\een{\end{enumerate}}
\def\bit{\begin{itemize}}
\def\eit{\end{itemize}}
\def\bei{\begin{itemize}}
\def\eei{\end{itemize}}
\def\bet{\begin{tabular}}
\def\eet{\end{tabular}}
\def\span{\text{span}}

\newcommand{\allcaps}[1]{\uppercase\expandafter{#1}}

\providecommand{\norm}[1]{\left\|#1\right\|}

\theoremstyle{definition}

\newtheorem{lemma}{Lemma} 
\newtheorem{remark}{Remark}

\makeatletter
\let\OldStatex\Statex
\renewcommand{\Statex}[1][3]{%
  \setlength\@tempdima{\algorithmicindent}%
  \OldStatex\hskip\dimexpr#1\@tempdima\relax}
\makeatother

\allowdisplaybreaks
\title{\LARGE \bf
Feature-Based Echo-State Networks: A Step Towards Interpretability and Minimalism in Reservoir Computer
}

\author{Debdipta Goswami
\thanks{Debdipta Goswami is an Assistant Professor in the Department of Mechanical and Aerospace Engineering, The Ohio State University, Columbus, OH 43210.
        {\tt\small goswami.78@osu.edu}}%
}

\begin{document}
\maketitle

\tikzstyle{block} = [draw, fill=blue!20, rectangle, 
    minimum height=3em, minimum width=4em]
\tikzstyle{sum} = [draw, fill=blue!20, circle, node distance=1cm]
\tikzstyle{input} = [coordinate]
\tikzstyle{noise} = [coordinate]
\tikzstyle{output} = [coordinate]
\tikzstyle{pinstyle} = [pin edge={to-,thin,black}]

\thispagestyle{empty}
\pagestyle{empty}

\begin{abstract}
This paper proposes a novel and interpretable recurrent neural-network structure using the echo-state network (ESN) paradigm for time-series prediction. While the traditional ESNs perform well for dynamical systems prediction, it needs a large dynamic reservoir with increased computational complexity. It also lacks interpretability to discern contributions from different input combinations to the output. Here, a systematic reservoir architecture is developed using smaller parallel reservoirs driven by different input combinations, known as features, and then they are nonlinearly combined to produce the output. The resultant feature-based ESN (Feat-ESN) outperforms the traditional single-reservoir ESN with less reservoir nodes. The predictive capability of the proposed architecture is demonstrated on three systems: two synthetic datasets from chaotic dynamical systems and a set of real-time traffic data. 


\end{abstract}

\section{INTRODUCTION}

Recent developments in machine-learning techniques for modeling and forecast of complex systems have become useful in a wide variety of problems, e.g., classification, speech recognition \cite{Hinton2012}, board games \cite{Silver2016}, and even discovering mathematical algorithms \cite{Fawzi2022}. Recurrent neural networks (RNNs) have been particularly useful for model-free prediction of dynamical systems. For example, an echo-state network (ESN) can model a chaotic system quite effectively \cite{Lu2017}, \cite{Pathak2018}. However, an ESN uses a relatively large reservoir of randomly connected nonlinear neurons to encode the dynamics from input-output data that can be computationally challenging for high-dimensional systems and lacks interpretability.

Neural network predictors, instead of using a physics-based handcrafted dynamic model, utilize the rich training dataset to build a parametric surrogate model, and then use it to predict the system outputs. An ESN is a special type of RNN that uses a reservoir of nonlinear, randomly connected neurons to process time-varying input signal. Such a network with a convergence property, known to the ESN literature as echo-state property (ESP), can uniformly approximate any nonlinear fading memory filter \cite{Ortega2018}. The ESN is attractive as a neural model since it can be trained via output connections with least-square method, thereby removing the need for back-propagation through time (BPTT) and saving computing resources. Also, a reservoir can be directly implemented by hardwares using field programmable gate arrays (FPGAs) or a photonic reservoir, thereby increasing efficiency and reducing computational overhead \cite{Tanaka2019}, \cite{Nakajima}. It is also extended to quantum computing realm via quantum reservoir computers (QRCs) \cite{Fuji2017}. ESN-based approaches are proved to be effective for sparse estimation of chaotic system and traffic volume on a road network \cite{Lu2017, Goswami2021, Goswami2022}. A training algorithm for noisy training dataset for an ESN is also developed via data-assimilation \cite{Goswami2023-2}.

This paper proposes a novel architecture for an ESN reservoir that excels in prediction with fewer reservoir nodes and provides interpretability to the prediction. It utilizes smaller combinations of input, termed as \emph{features} fed into separate smaller reservoirs independently. The output from all the parallel smaller reservoirs are then combined to produce the desired system output. The relative strength of each output weight from the respective reservoir provides an interpretable contribution of each feature to the system output. The resultant feature-based echo-state network (Feat-ESN) greatly reduces the number of reservoir nodes/neurons require for an effective prediction capability. This is particularly useful for high dimensional system, e.g., traffic volumes on a road network.

The contribution of this paper are (1) developing a novel ESN architecture with parallel smaller reservoirs to provide interpretability to the ESN approach; (2) significantly reducing the number of reservoir nodes necessary for effective prediction by interpretable choice of input features; (4) extension of the algorithm for partial measurements as the training data by delay-embedding; and (4) application of the prediction method on a real set of mobility data in order to forecast traffic volume in a road network.

This paper is organized as follows. Section II provides a brief overview of the echo-state network (ESN). Section III builds the novel architecture and presents the Feat-ESN algorithm. Section IV illustrates the applications to three different problems: two synthetic data streams generated by chaotic nonlinear systems and one real set of traffic sensor data. An ablation study with different block-size is also provided. Section V concludes the manuscript and discusses ongoing and future work.

\section{Echo-State Networks for Dynamical Systems Prediction}
Echo-state networks are special type of recurrent neural network consisting of a large dynamic reservoir of randomly connected neurons driven nonlinearly by input signals. These neuronal responses are then linearly combined to match a desired output signal. An ESN's performance is heavily dependent on the richness of the dynamic reservoir and hence, it is also called a reservoir computer (RC). An ESN consists of an input layer $\mf{u}\in\mb{R}^m$, coupled through input coupling matrix $W_{in}\in \mb{R}^{n\times m}$ with a recurrent nonlinear reservoir $\mf{r} \in \mathbb{R}^n$. The output $\mathbf{y}\in \mathbb{R}^p$ is generated from $n$ neurons of the reservoir via a readout matrix $W_{out}\in \mb{R}^{n\times p}$. The reservoir network evolves nonlinearly in following  fashion \cite{Maass2004}, \cite{Goswami2021}
\bql
\mf{r}(t+\Delta t)=(1-\alpha)\mf{r}(t) + \alpha\psi(W\mf{r}(t)+W_{in}\mf{u}(t) + \mf{d}).
\eql
The time-step $\Delta t$ denotes the sampling interval of the training data and $\mf{d}\in\mb{R}^n$ is a randomly chosen bias with elements between $(-0.5, 0.5)$. The leakage rate parameter $\alpha \in (0,1]$ helps slowing down the evolution of the reservoir states as $\alpha\rightarrow 0$. The nonlinear activation function $\psi(\cdot)$ is usually a sigmoid function, e.g., $\tanh(\cdot)$. The output $\mf{y}(t)$ is linearly read out from the reservoir states \cite{Maass2004}, \cite{Goswami2021}, i.e.,
\bql
\mf{y}(t)=W_{out}\mf{r}(t).
\eql
The weights $W_{in}$ and $W$ are initially randomly drawn and then held fixed. The weight $W_{out}$ is adjusted during the training process. The reservoir weight matrix $W$ is usually kept sparse for computational efficiency.

During the training phase, an ESN is driven by an input sequence $\{\mf{u}(t_1),\ldots,\mf{u}(t_N)\}$ that yields a sequence of reservoir states $\{\mf{r}(t_1),\ldots,\mf{r}(t_N)\}$. The reservoir states are stored in a matrix $\mf{R}=[\mf{r}(t_1),\ldots,\mf{r}(t_N)]$. The correct outputs $\{\mf{y}(t_1),\ldots,\mf{y}(t_N)\}$, which are part of the training data, are also arranged in a matrix $\mf{Y}=[\mf{y}(t_1),\ldots,\mf{y}(t_N)]$.  The training is carried out by a linear regression with Tikhonov regularization as follows \cite{Jaeger2004}:
\bql \label{Eq: LSTrain}
W_{out} = \mf{Y}\mf{R}^T(\mf{R}\mf{R}^T + \beta\mf{I})^{-1},
\eql
where $\beta>0$ is a regularization parameter to ensure non-singularity. 
\begin{figure*}[t]
\centering 
\subfloat[]{\includegraphics[trim=0cm 0cm 0cm 0cm, clip=true, width=0.5\textwidth]{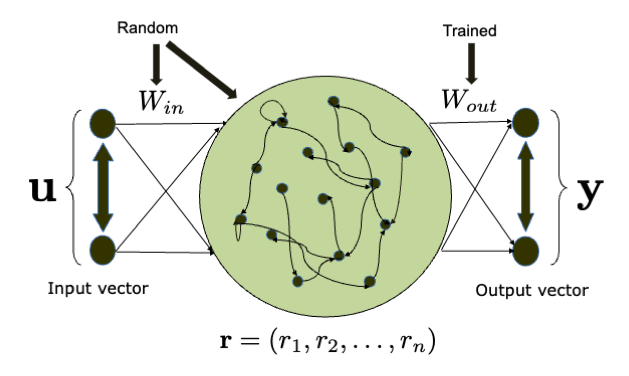}}\hspace{0cm}
\subfloat[]{\includegraphics[trim=1cm 2cm 1cm 2cm, clip=true, width=0.4\textwidth]{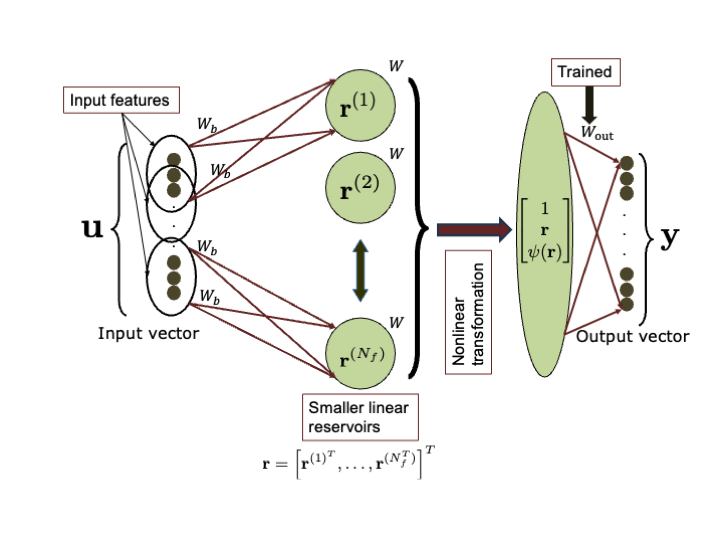}}
\caption{Architecture and training of an ESN: (a) the basic ESN, (b)  Feature-based ESN (Feat-ESN) } \label{Fig: Reservoir}
\end{figure*}

\begin{remark}
Reference \cite{Ortega2018} shows that an ESN is a universal approximator, i.e., it can realize any nonlinear operator with bounded memory arbitrarily accurately if it satisfes the echo-state property (ESP) \cite{Jaeger2004}. An ESN is said to have the ESP if the reservoir asymptotically washes out any information from the initial conditions. For the $\tanh(\cdot)$ activation function, it is empirically observed that the ESP holds for any input if the spectral radius of $W$ is smaller than unity \cite{Jaeger2004}. To ensure this condition, $W$ is normalized by its spectral radius. 
\end{remark}

\section{Feature-Based Echo-State Network: A Bite Sized Approach}
An ESN provides a great predictive model for a time series $\{\mf{x}(t_i)\in \mb{R}^d:i\in\mb{N}\}$ generated by a dynamical system by setting $\mf{u}(t)$ and $\mf{y}(t)$ as the current and next state value (i.e., $\mf{x}(t_k)$ and $\mf{x}(t_{k+1})$) respectively. The network is trained for a certain training length $N$ of the time-series data $\{\mf{x}(t_i,i=1,\ldots,N\}$, which can then run freely by feeding the output $\mf{y}(t_k)$ back to the input $\mf{u}(t_{k+1})$ of the reservoir. In this case, both $\mf{u}$ and $\mf{y}$ have the same dimension $d$ as that of the time-series data. This setup is shown in Fig.~\ref{Fig: Reservoir}(a). 

An ESN proves to be a powerful tool for dynamical systems prediction when trained with noiseless data \cite{Lu2017}, \cite{Pathak2018}. Its performance is significantly improved when partial observations are available during the testing phase by assimilating them through an ensemble Kalman filter \cite{Goswami2021}. It can also be modified to accommodate partial state measurements as training data by a higher dimensional delay-embedding in the input layer \cite{Goswami2023}. However, due to the inherent randomness of the reservoir connections, relatively large number of reservoir nodes are required for any meaningful prediction performance. Moreover, it is not possible to infer which reservoir-node has more contribution to the input-output prediction performance, thereby lacking interpretability.

This paper proposes a systematic approach to generate a reservoir using input combination as \emph{features} and corresponding smaller reservoirs as \emph{feature maps} in order to reduce the overall reservoir size. 
The proposed architecture has three components:
\paragraph{\textbf{Input map}} A collection of $N_f$ features is selected from the input vector $\mf{u} = \{u_1,\ldots u_m\} \in \mb{R}^m$. A feature is any combination of the input components, i.e., $\{u_{i_1},\ldots, u_{i_l}\}$ where $l$ can be between $1$ and $m$. For an $m$-dimensional input, maximum $2^m-1$ features can be extracted. The input matrix $W_{in}$ is such chosen that each feature is fed separately into smaller individual reservoirs. To construct the input matrix, a feature matrix $[W_f\in \mb{R}^{N_f\times m}]$ is constructed such that 
\bql\label{Eq: Feature_matrix}
W_{f_{ij}}=\begin{cases}
1 & \text{if $u_j$ is in $i^{\text{th}}$ feature}\\
0 & \text{if $u_j$ is not in $i^{\text{th}}$ feature}.
\end{cases}
\eql

\begin{algorithm}
\footnotesize
\caption{Feat-ESN: Realization and Training of a feature-based ESN}\label{Alg: FeatESN}
\textbf{Input:} Training input data $\{\mf{u}(t_1),\ldots,\mf{u}(t_N)\}$, $\mf{u}(t_i) \in \mb{R}^m$, training output data $\{\mf{y}(t_1),\ldots,\mf{y}(t_N)\}$, $\mf{y}(t_i) \in \mb{R}^p$\\
\textbf{Hyperparameters:} Training length $N$, leaking rate $\alpha$, regularization parameter $\beta$, reservoir connection probability $p\in(0,1)$, feature matrix $W_f$, block size $b$, nonlinear readout $\psi$\\
\textbf{Output:} $W_{in}$, $W$, $W_{out}$
\begin{algorithmic}[1]
\Procedure {Train}{ $\{\mf{y}(t_1),\ldots,\mf{y}(t_N)\}$; $\alpha$, $\beta$, $p$, $n$, $\psi$, $M$, $\Sigma_{x}$, $\Sigma_w$, $\Sigma_v$}

\State Generate $W_b\in \mb{R}^{b}\sim \operatorname{unif}(-0.5, 0.5)^b$ random vector
\State Generate $W_r\sim G(b,p)$ \Comment{Adjacency matrix of an Erd\"os-Renyi random graph}
\State Compute $W_{in} = W_f \otimes W_b$
\State Compute $W_{in} = I_{N_f} \otimes W_r$
\State $\mf{Y} \gets [\mf{y}(t_1),\ldots,\mf{y}(t_N)]$ \Comment{Arrange outputs}
\State $\mf{r}_{t_1} \gets \mf{0}_{N_fb}$ \Comment{Initialize reservoir}
\For{$k=1$ to $N$}
\State $\mf{r}(t_{k+1}) \gets (1-\alpha)\mf{r}(t_{k}) $ \Statex $+ \alpha\left(W\mf{r}(t_{k})+W_{in}\mf{u}(t_{k})+\mf{d}\right)$
\State $\Psi(\mf{r}(t_k)) = [1 \quad \mf{r}(t_k)^T\quad \psi(\mf{r}(t_k))^T]^T$
\EndFor
\State $\mf{\Psi} \gets [\Psi(\mf{r}(t_1)),\ldots,\Psi(\mf{r}(t_N))]$ \Comment{Arrange nonlinear reservoir readouts}
\State $W_{out} \gets \mf{Y}\mf{\Psi}^T(\mf{\Psi}\mf{\Psi}^T + \beta\mf{I})^{-1}$ \Comment{Train output weights}
\EndProcedure
\end{algorithmic}
\end{algorithm}

For each such feature, a random block-vector $W_b \in \mb{R}^b$ of block-size $b\in\mb{N}$ is generated uniformly between $-0.5$ and $0.5$ and the input matrix is given by
\bql \label{Eq: Input_matrix}
W_{in}(\in \mb{R}^{N_fb\times m}) = W_f\otimes W_b = \begin{bmatrix} W_{in_1}\\\vdots \\W_{in_{N_f}}\end{bmatrix}.\eql  For a three-dimensional input with all possible features (i.e., $N_f = 7$), the input matrix becomes
\[ \footnotesize W_{in} = 
\left .\begin{bmatrix} W_b & 0 & 0 \\
                0 & W_b & 0 \\
                0 & 0 & W_b \\
                W_b & W_b & 0 \\
                W_b & 0 & W_b \\
                0 & W_b & W_b \\
                W_b & W_b & W_b
                \end{bmatrix}\right._{7b\times 3}.\]
                
\paragraph{\textbf{Reservoirs}} Each feature is fed into separate smaller reservoirs $\mf{r}^{(i)} \in \mb{R}^b,\quad,i\in\{1,\ldots,N_f\}$ such that the reservoir state is $\mf{r} = \left[\mf{r}^{(1)^T},\ldots, \mf{r}^{(N_f)^T}\right]^T \in \mb{R}^{N_fb}$. Each reservoir evolves linearly in the following fashion
\bql \label{Eq: novel_reservoir}
\mf{r}^{(i)}(t_{k+1}) = (1-\alpha)\mf{r}^{(i)}(t_{k}) + \alpha(W_r\mf{r}^{(i)}(t_{k})+W_{in_i}\mf{u}(t) + \mf{d}^{(i)}),
\eql
where $\alpha$ is the leaking rate, $W_r\in \mb{R}^{b\times b}$ is the reservoir transition matrix, and $\mf{d}^{(i)}\in\mb{R}^{b}$ is the bias. Matrix $W_r$ is randomly drawn from a sparse random-graph model, e.g., Erd\"{o}s-Renyi model and normalized by a desired spectral radius in order to maintain the echo-state property. Hence, the total number of reservoir nodes is $n=N_fb$ and the overall reservoir transition matrix $W = I_{N_f} \otimes W_r$ with a linear reservoir dynamics $\mf{r}(t_{k+1}) = (1-\alpha)\mf{r}(t_{k}) + \alpha\left(W\mf{r}(t_{k})+W_{in}\mf{u}(t) + \mf{d}\right)$. Each smaller reservoir $\mf{r}^{(i)}$ defines feature-map that depends on the $i^{\text{th}}$ feature only.

\paragraph{\textbf{Readout}} To maintain the expressivity of the echo-state network with linear reservoir and fewer reservoir nodes, a nonlinear readout network is used. To maintain the ease of least-squares training, we take a combination of the reservoir, its nonlinear map, and a bias $\Psi(\mf{r})\triangleq\left[1 \quad \mf{r}^T \quad \psi(\mf{r})^T\right]^T$ with $\psi(\cdot):\mf{R}^{n} \rightarrow \mf{R}^q$. The output is then linearly read out from $\Psi(\mf{r})$ is $\mf{y}(t_k)=W_{out}\Psi(\mf{r}(t_k))$. Choice of the nonlinearity $\psi(\cdot)$ and its rank $q$ are hyperparameters. 

\begin{remark}
Each smaller reservoir $\mf{r}^{(i)}$ is forced with only the corresponding feature and evolves independently with other ones. The magnitude of output weights $W_{out}^{(i)}$ associated with each reservoir provides a metric of contribution of each feature to the output. 
\end{remark}

\begin{remark}
The relative magnitudes of $W_{out}^{(i)}$ can be utilized to prune the reservoir even more by removing the reservoir $\mf{r}^{(i)}$ with $\norm{W_{out}^{(i)}}$ smaller than a predefined threshold. 
\end{remark}

Feat-ESN with a suitable nonlinear readout also satisfies the universal approximation property. The following lemma explains it.
\begin{lemma}
If the feature matrix $W_f$ has the full column rank, $m<N_fb$, and the readout functions are chosen from a subalgebra of $C(\mb{R}^n,\mb{R}^n)$ that separate points, then the Feat-ESN maps $\{\mf{u}(t_k)\}_k \mapsto \{\mf{y}(t_k)\}_k$ are dense in $C(\mb{U},\mb{Y})$ where $\mb{U}$ and $\mb{Y}$ denotes the spaces of $\mb{R}^m$ and $\mb{R}^p$ valued sequences respectively.
\end{lemma}

\begin{proof}
Since $W_f$ is full rank and $m<N_fb$, $W_{in}$ is full rank, i.e., the map $\{\mf{u}(t_k)\}_k\mapsto \{\mf{r}(t_k)\}_k$ separates points. From the hypothesis of the lemma, the readouts are dense and separate points as well. From the linear construction of the reservoir, the maps $\{\mf{u}(t_k)\}_k \mapsto \{\mf{y}(t_k)\}_k$ form a subalgebra of $C(\mb{U},\mb{Y})$.  Application of Stone-Weierstrass theorem for locally compact Hausdorff $\mb{U}$ yields that the aforemntioned maps are indeed dense in $C(\mb{U},\mb{Y})$.
\end{proof}

\section{Numerical Examples}
This section illustrates the performance and ablation study of the Feat-ESN algorithm on three time series data. The first two are time-series generated by chaotic dynamical systems and the last one is a real-time traffic flow data obtained by Numina sensor nodes \cite{Numina} installed on the University of Maryland campus. To make a fair comparison between a regular ESN and Feat-ESN we use the same number of reservoir nodes $n=N_fb$ in the regular ESN. The Feat-ESN achieves better accuracy with very small number of reservoir nodes as depicted in the results.

\subsection{Lorenz System}

The Feat-ESN algorithm is tested on a time-series $[x(t_k)\,\, y(t_k) \,\, z(t_k)]$ generated by the Lorenz system:
\bnl\label{Eq: Lorenz}
\dot{x} &=& \sigma(y-x)\\\nonumber
\dot{y} &=& x(\rho-z) -y \\\nonumber
\dot{z} &=& xy - \beta z,
\enl
\begin{figure}[t]
\centering 
\subfloat[]{\includegraphics[trim=0cm 0cm 0cm 0cm, clip=true, width=0.25\textwidth]{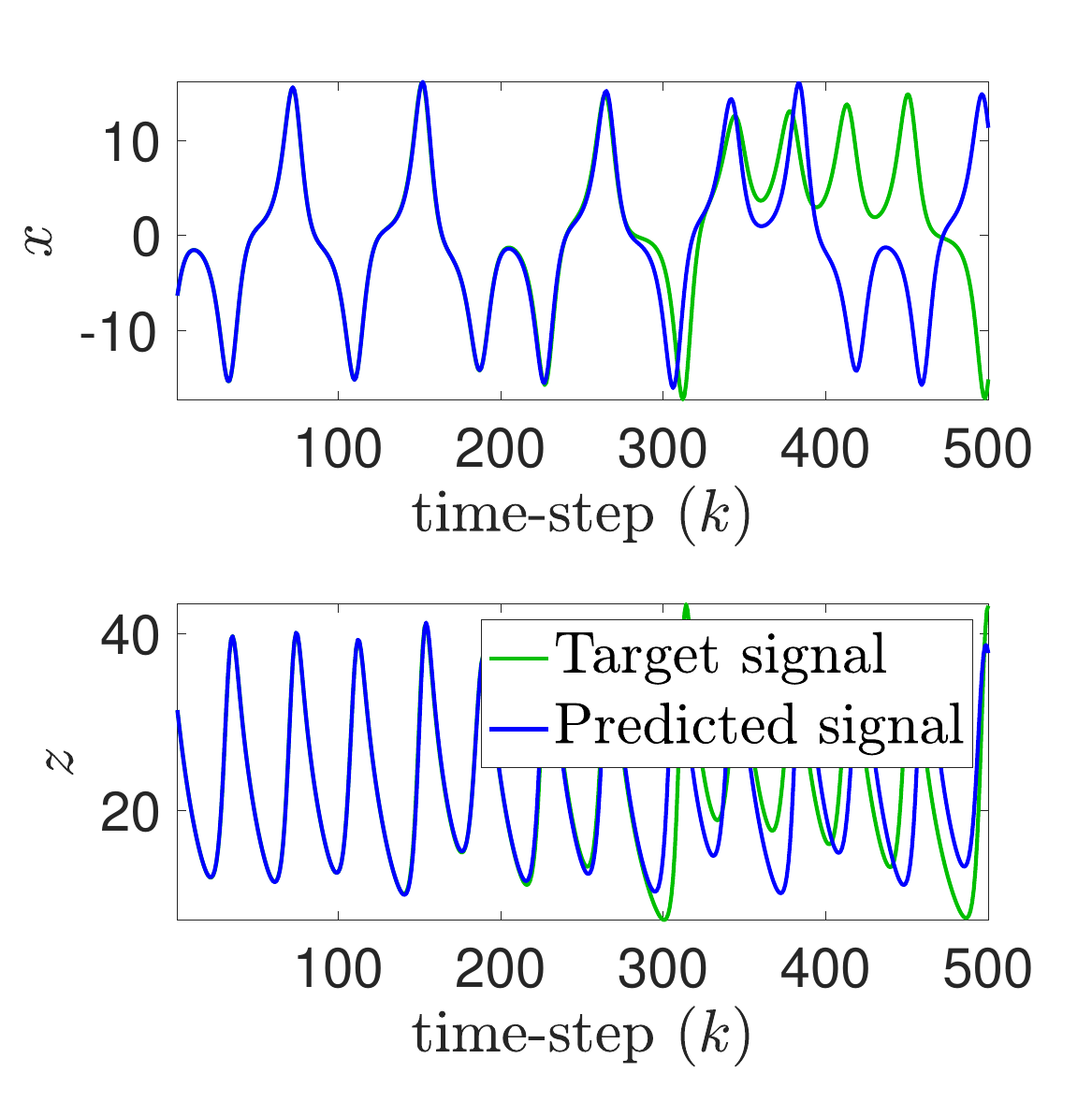}}
\subfloat[]{\includegraphics[trim=0cm 0cm 0cm 0cm, clip=true, width=0.25\textwidth]{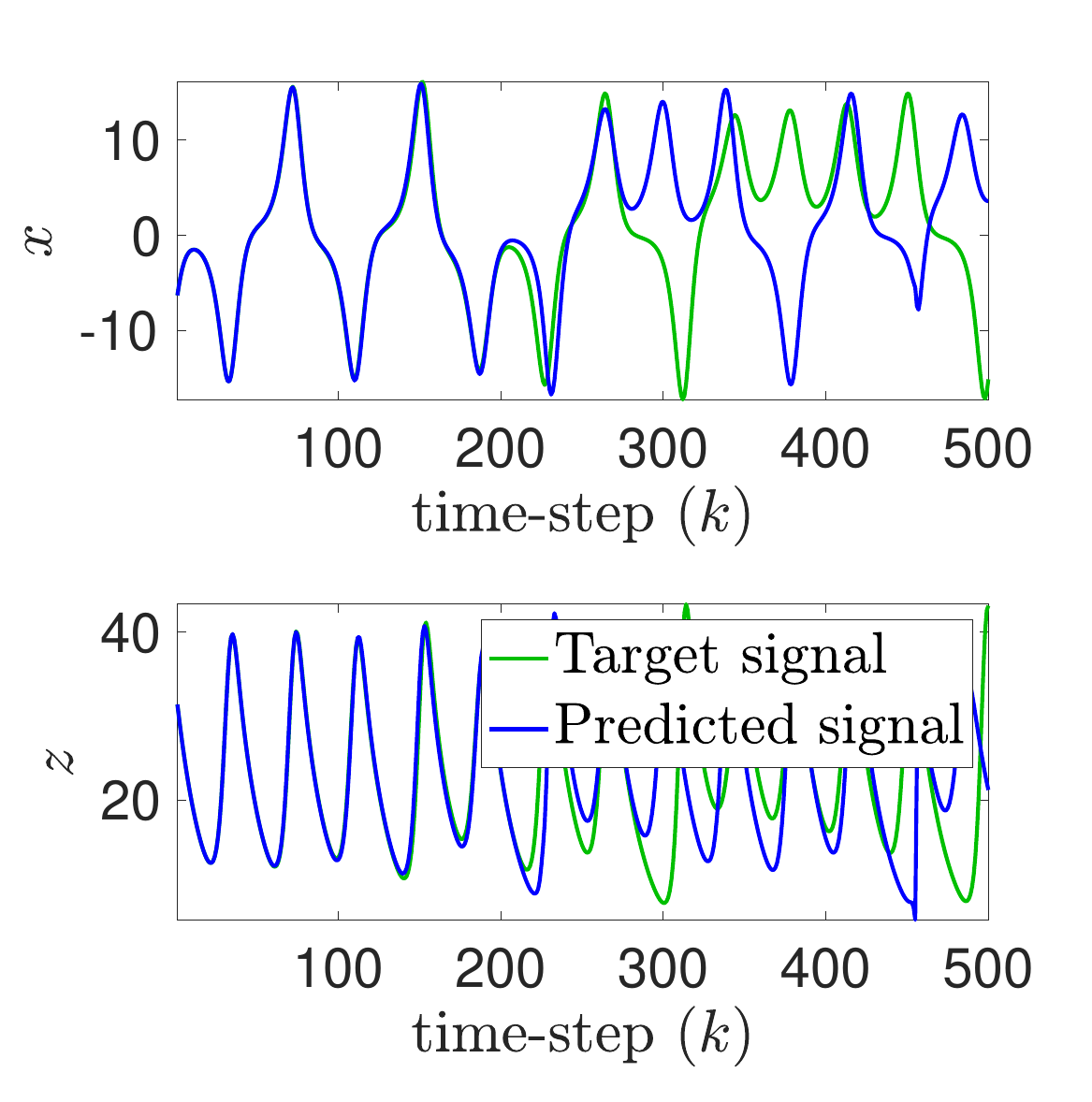}}
\caption{prediction of the noisy time-series $x(t_k)$ and $z(t_k)$ from Lorenz system \eqref{Eq: Lorenz} with $b=100$ and $N_f=7$, i.e., reservoir size $n=700$: (a) true and predicted signal with Feat-ESN, (b)  true and predicted signal with regular ESN} \label{Fig: LorenzPrediction}
\end{figure}
\begin{figure}[t]
\centering 
\includegraphics[trim=0cm 0cm 0cm 0cm, clip=true, width=0.5\textwidth]{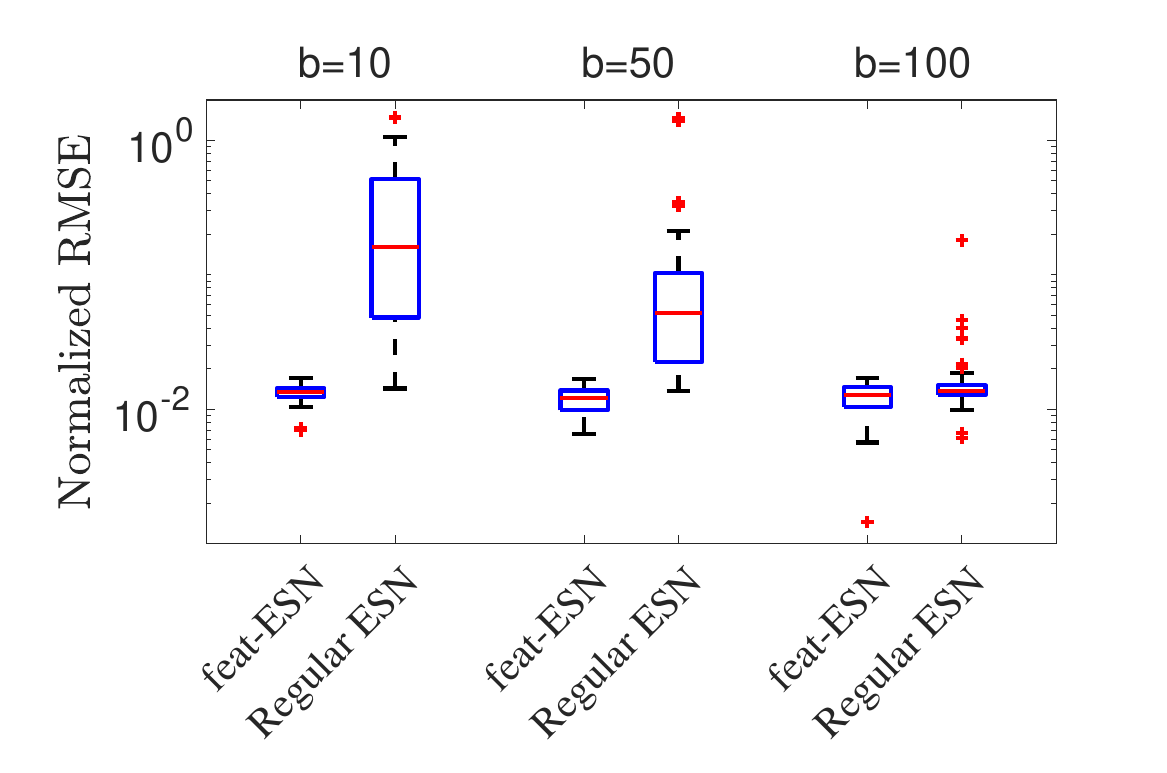}
\caption{Error profile of Lorenz time-series prediction: NRMSE with different block-size $b$} \label{Fig: LorenzError}
\end{figure}
\begin{figure}[!t]
\centering 
\includegraphics[trim=0cm 0cm 0cm 0cm, clip=true, width=0.5\textwidth]{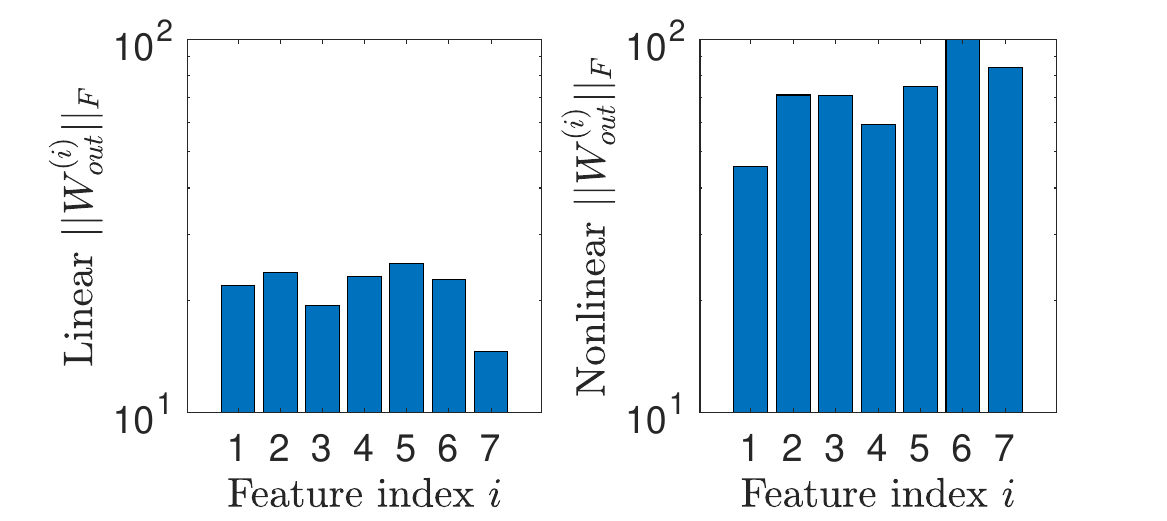}
\caption{Frobenius norm of the output map for different features of Lorenz time-series prediction} \label{Fig: LorenzOutput}
\end{figure}
\begin{figure}[t]
\centering 
\subfloat[]{\includegraphics[trim=0cm 0cm 0cm 0cm, clip=true, width=0.25\textwidth]{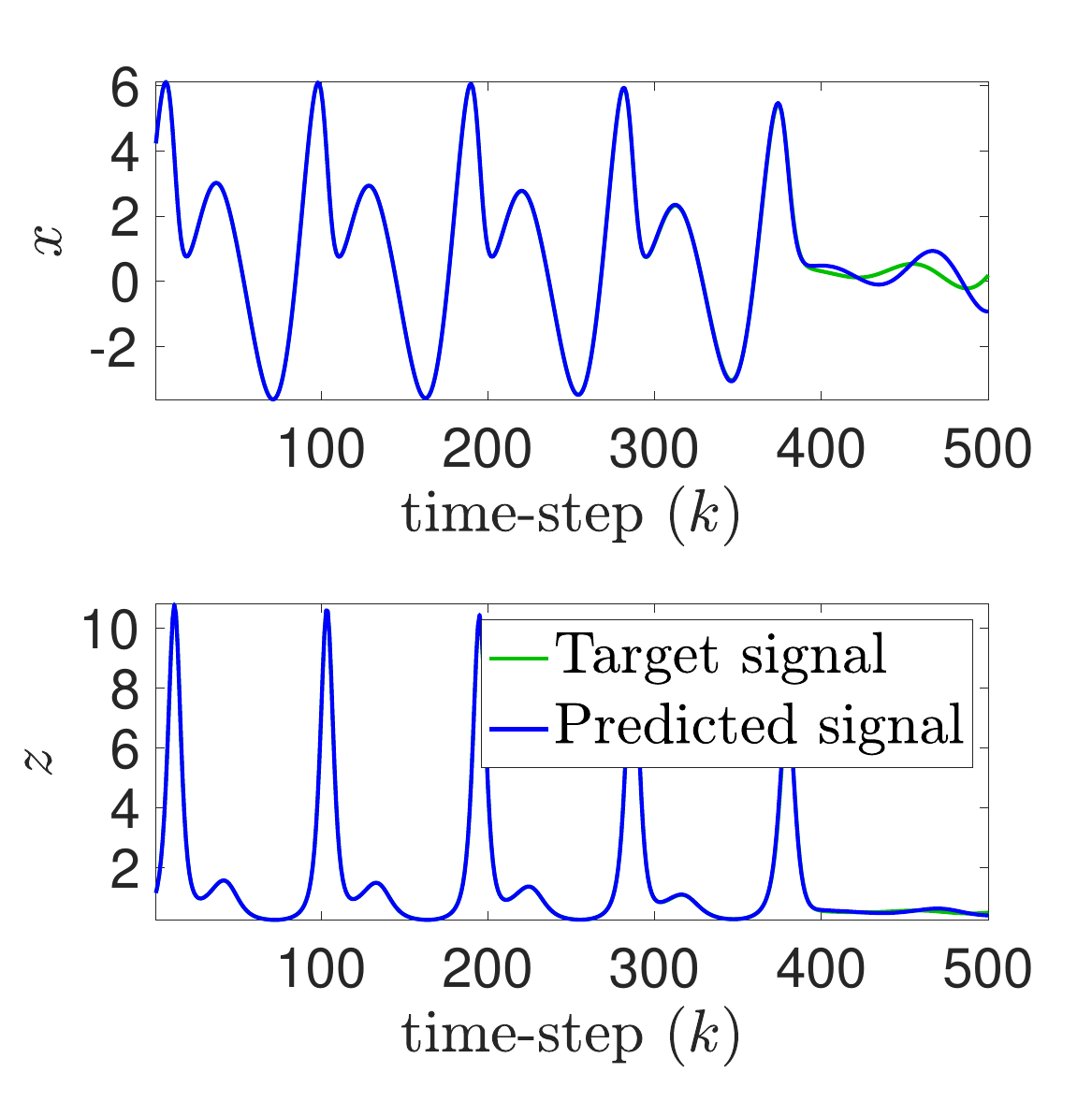}}
\subfloat[]{\includegraphics[trim=0cm 0cm 0cm 0cm, clip=true, width=0.25\textwidth]{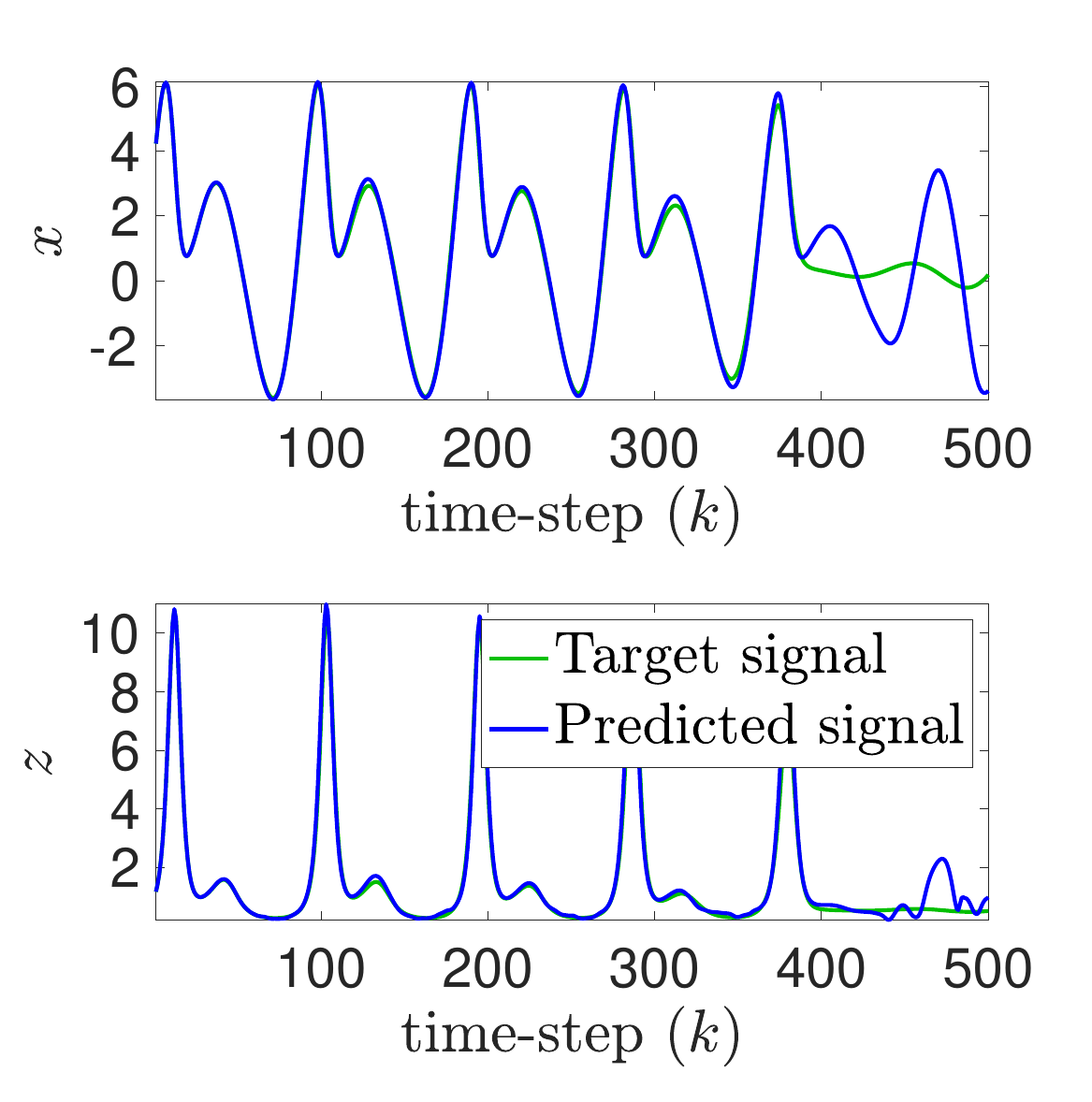}}
\caption{prediction of the noisy time-series $x(t_k)$ and $z(t_k)$ from R\"ossler system \eqref{Eq: Rossler} with $b=100$ and $N_f=7$, i.e., reservoir size $n=700$: (a) true and predicted signal with Feat-ESN, (b)  true and predicted signal with least square training} \label{Fig: RosslerPrediction}
\end{figure}
\begin{figure}[t]
\centering 
\includegraphics[trim=0cm 0cm 0cm 0cm, clip=true, width=0.5\textwidth]{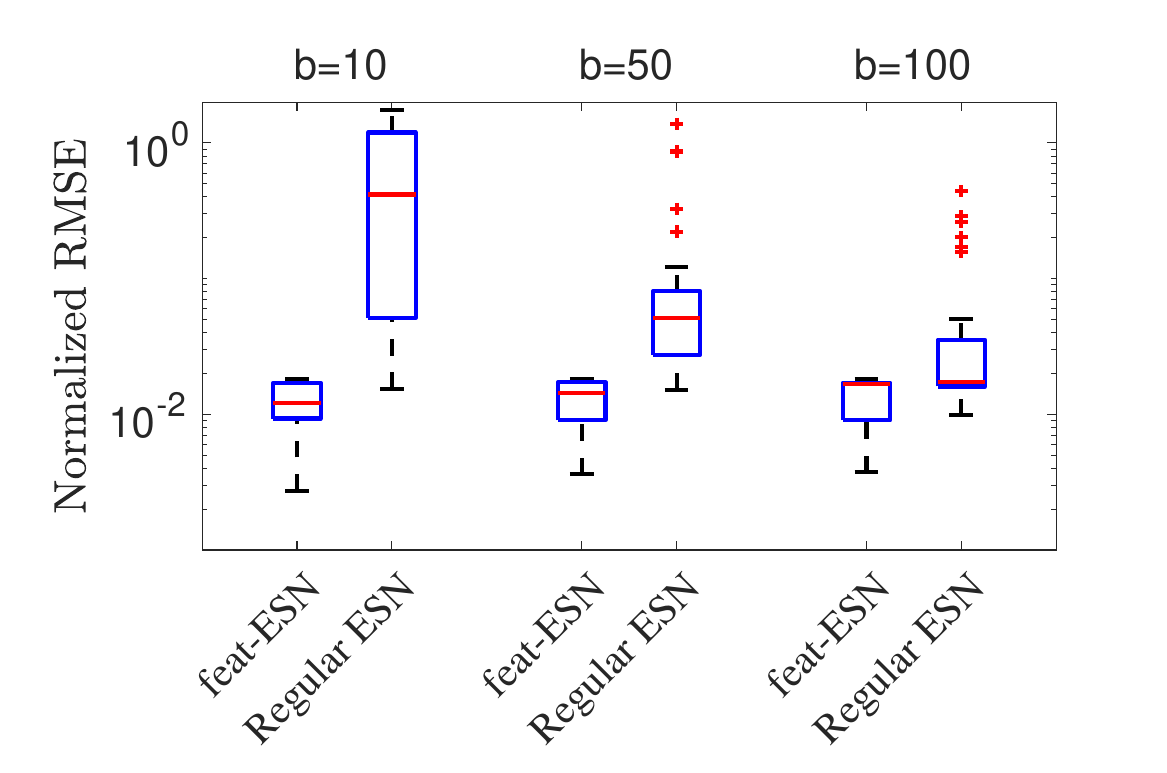}
\caption{Error profile of R\"ossler time-series prediction: NRMSE with different block-size $b$} \label{Fig: RosslerError}
\end{figure}
\begin{figure}[!t]
\centering 
\includegraphics[trim=0cm 0cm 0cm 0cm, clip=true, width=0.5\textwidth]{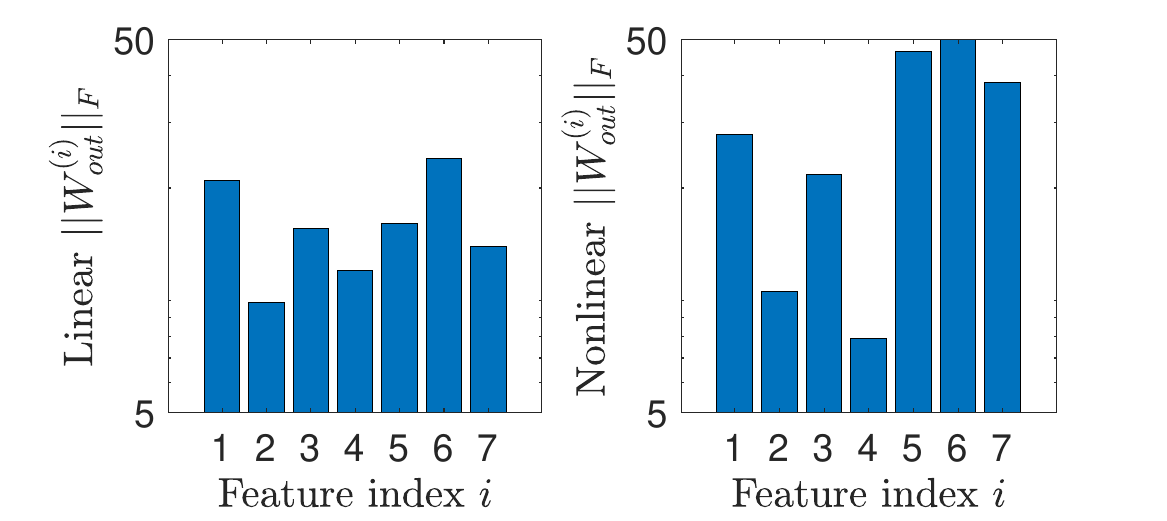}
\caption{Frobenius norm of the output map for different features of R\"ossler time-series prediction} \label{Fig: RosslerOutput}
\end{figure}
\begin{figure}[t]
\centering 
\subfloat[]{\includegraphics[trim=1cm 0.5cm 2cm 0.2cm, clip=true, width=0.25\textwidth]{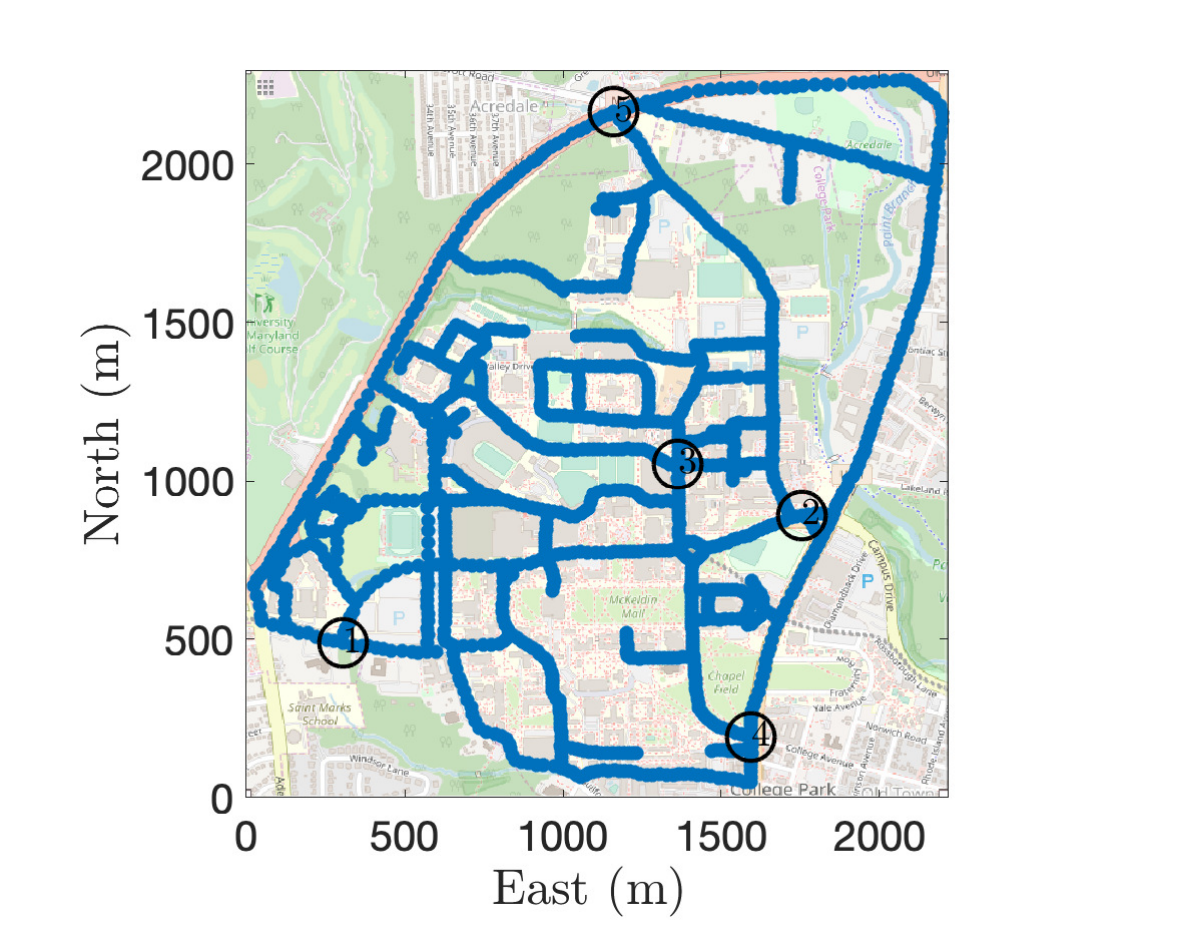}}
\subfloat[]{\includegraphics[trim=2cm 0cm 0cm 0cm, clip=true, width=0.25\textwidth]{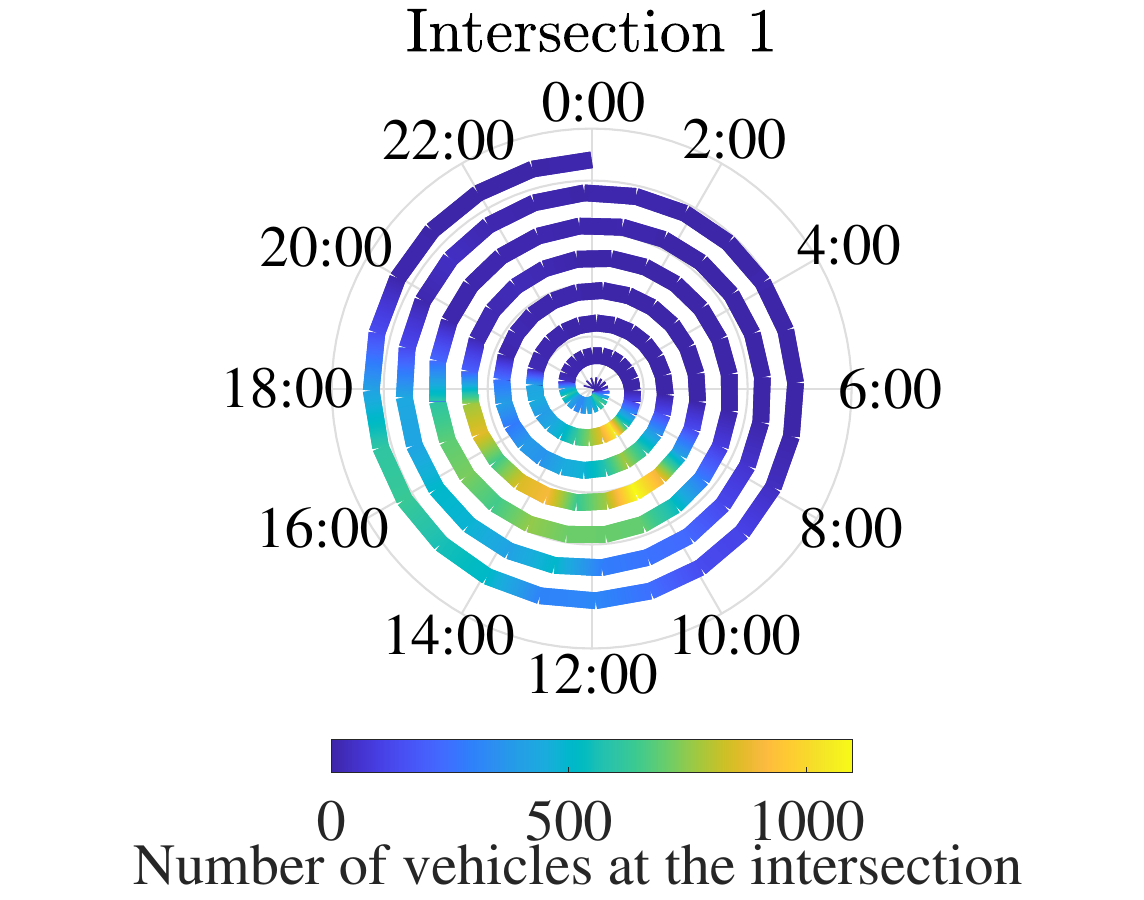}}
\caption{Schematic diagram of traffic data: (a) University of Maryland road network with Numina sensors, (b) Traffic congestion pattern of an intersection over a single week, each revolution denotes a day of the week with times marked as angles; the number of vehicles is denoted by the colormap.} \label{Fig: TrafficSchematic}
\end{figure}
\begin{figure}[!t]
\centering 
\includegraphics[trim=0cm 1cm 0cm 1cm, clip=true, width=0.4\textwidth]{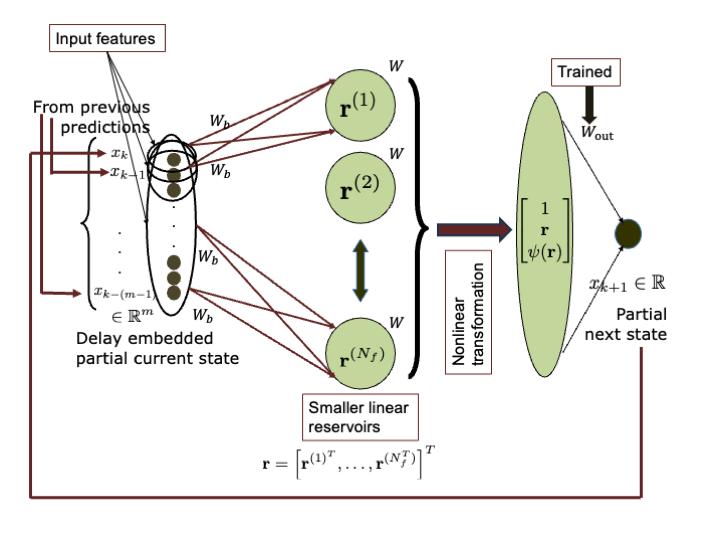}
\caption{$m$-dimensional delay-embedding of the scalar partial observation for Feat-ESN} \label{Fig: DelayFeatESN}
\end{figure}
\begin{figure}[t]
\centering 
\subfloat[]{\includegraphics[trim=0cm 0cm 0cm 0cm, clip=true, width=0.25\textwidth]{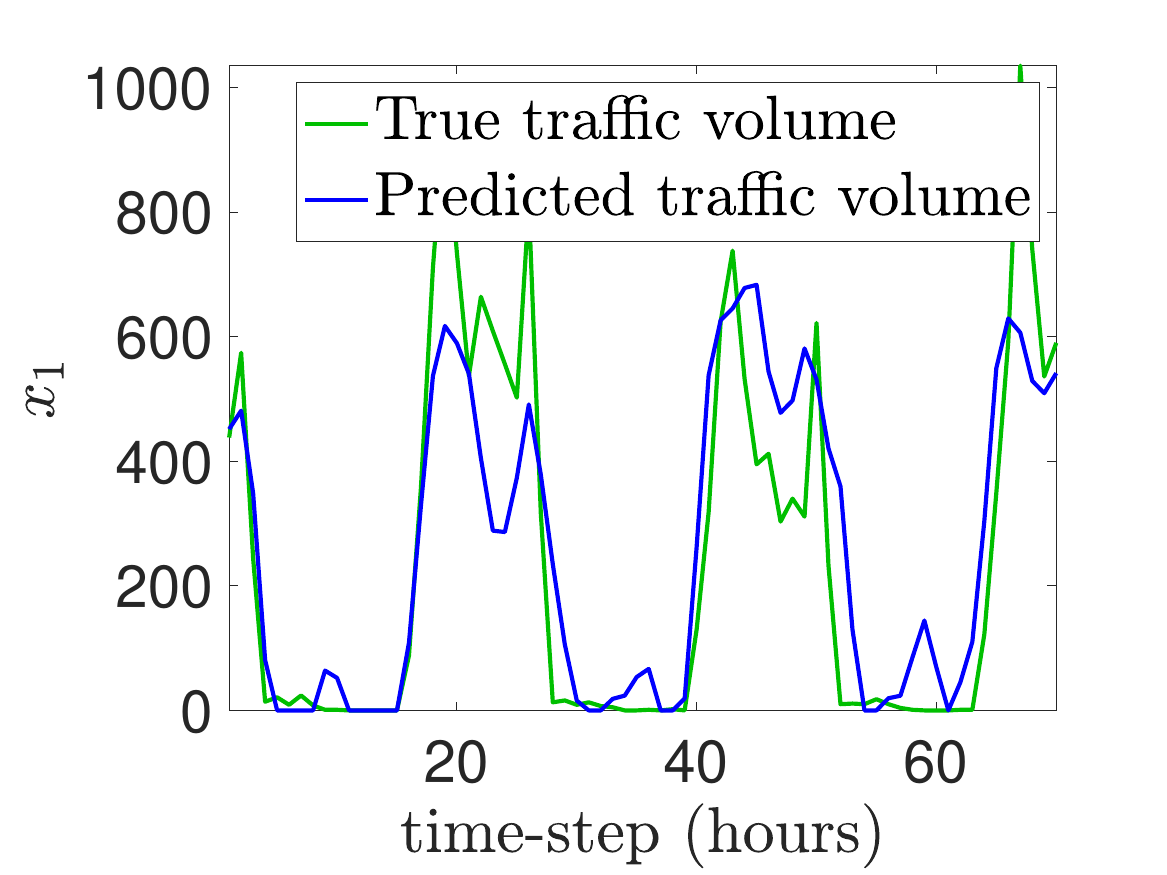}}
\subfloat[]{\includegraphics[trim=0cm 0cm 0cm 0cm, clip=true, width=0.25\textwidth]{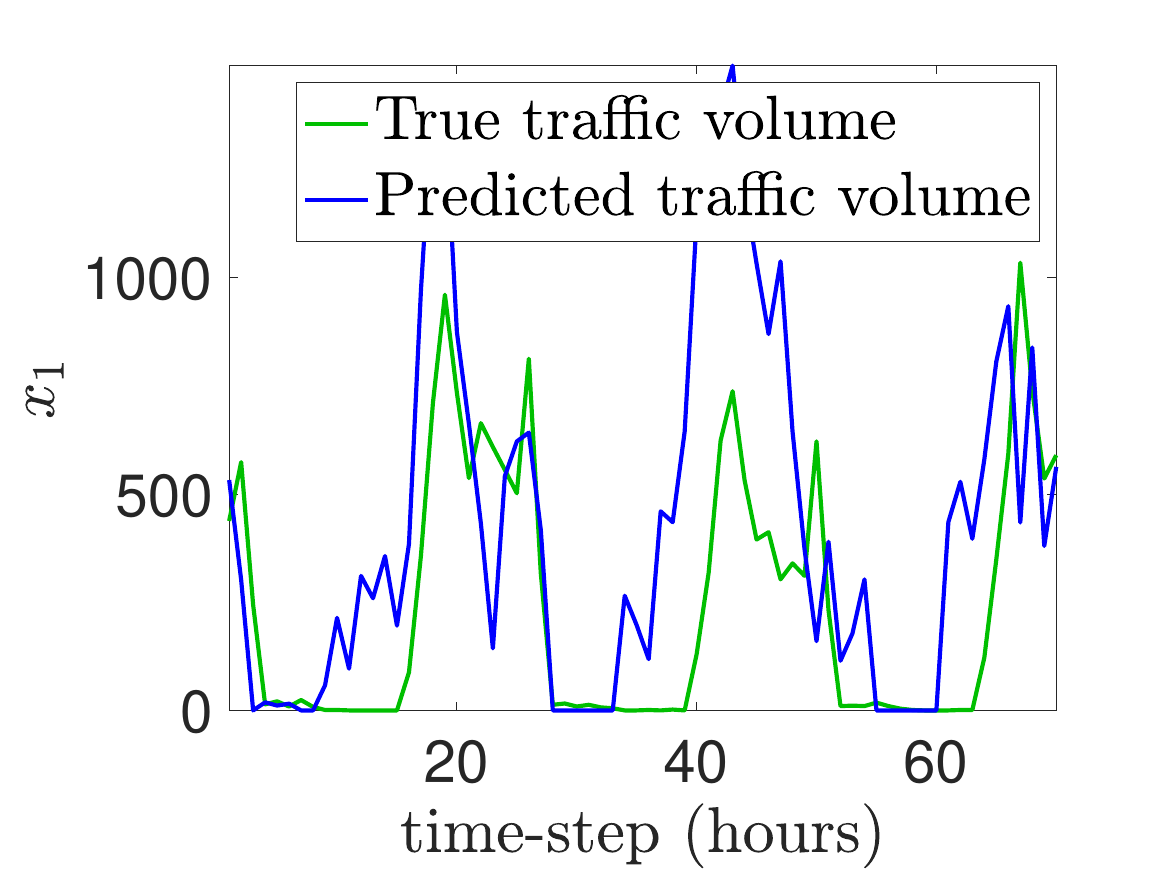}}
\caption{prediction of the time-series of traffic volume recorded in Numina sensor 1 with $b=10$ and $N_f=100$, i.e., reservoir size $n=1000$: (a) true and predicted traffic volume with Feat-ESN, (b)  true and predicted traffic volume with least square training} \label{Fig: NuminaPrediction}
\end{figure}
\begin{figure}[t]
\centering 
\subfloat[]{\includegraphics[trim=0cm 0cm 0cm 0cm, clip=true, width=0.25\textwidth]{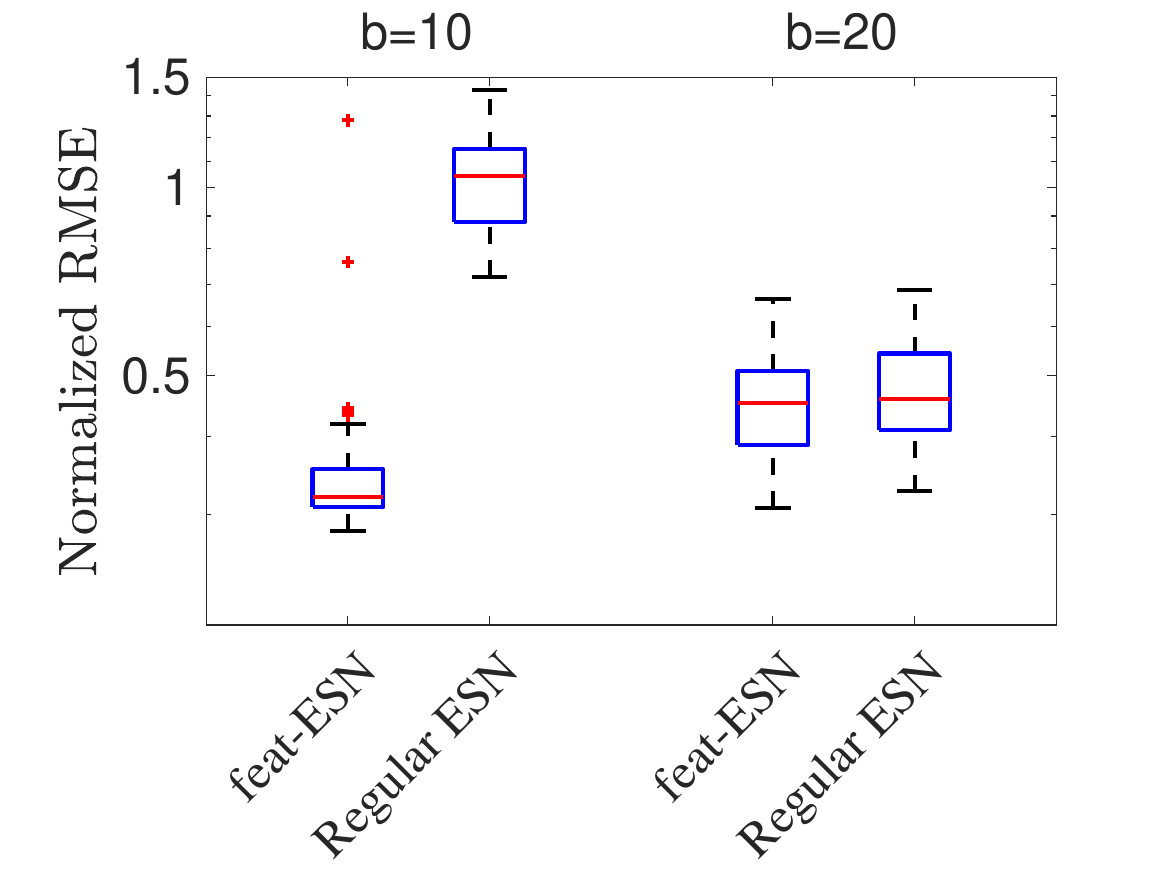}}
\subfloat[]{\includegraphics[trim=0cm 0cm 0cm 0cm, clip=true, width=0.25\textwidth]{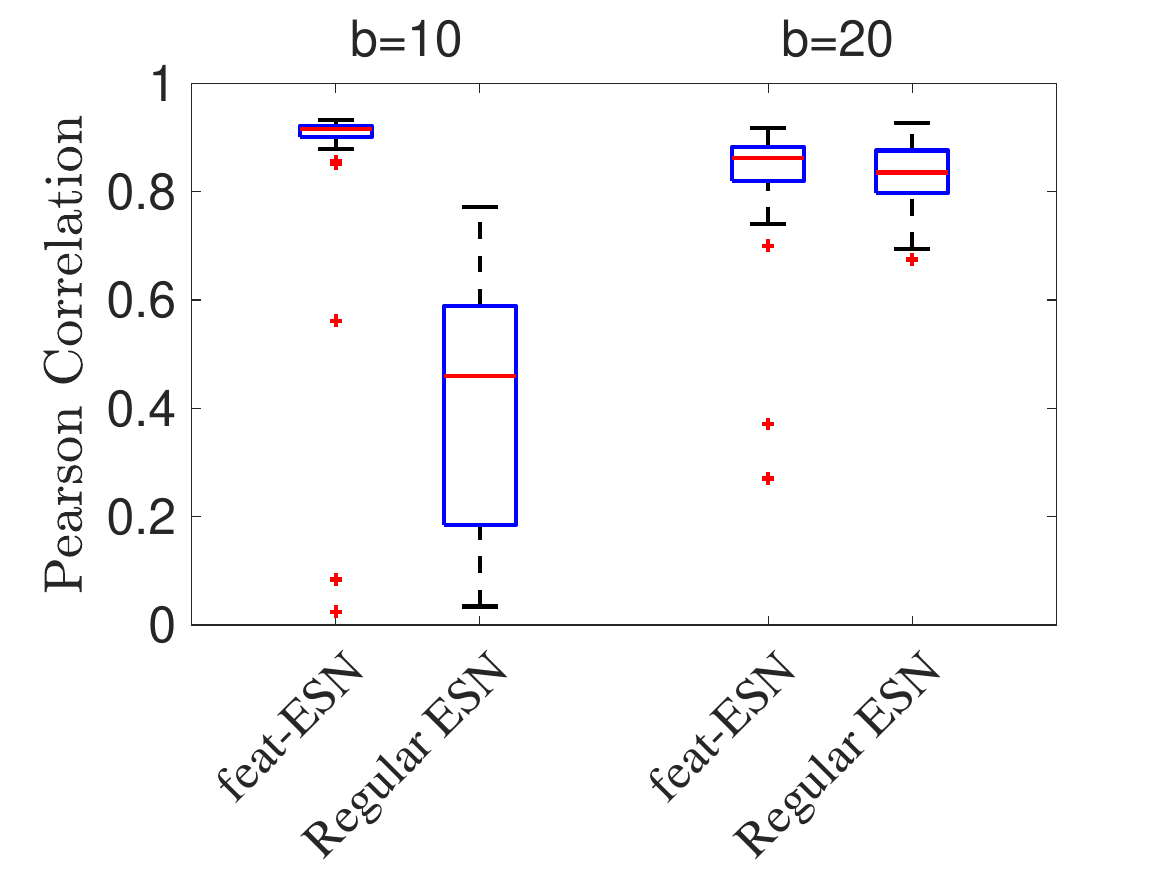}}
\caption{Error and correlation profile of traffic volume prediction: (a) NRMSE  and (b) Pearson correlation with different block-size $b$. The delay embedding dimension is $m=100$.} \label{Fig: NuminaError}
\end{figure}
\begin{figure}[!t]
\centering 
\includegraphics[trim=0cm 0cm 0cm 0cm, clip=true, width=0.5\textwidth]{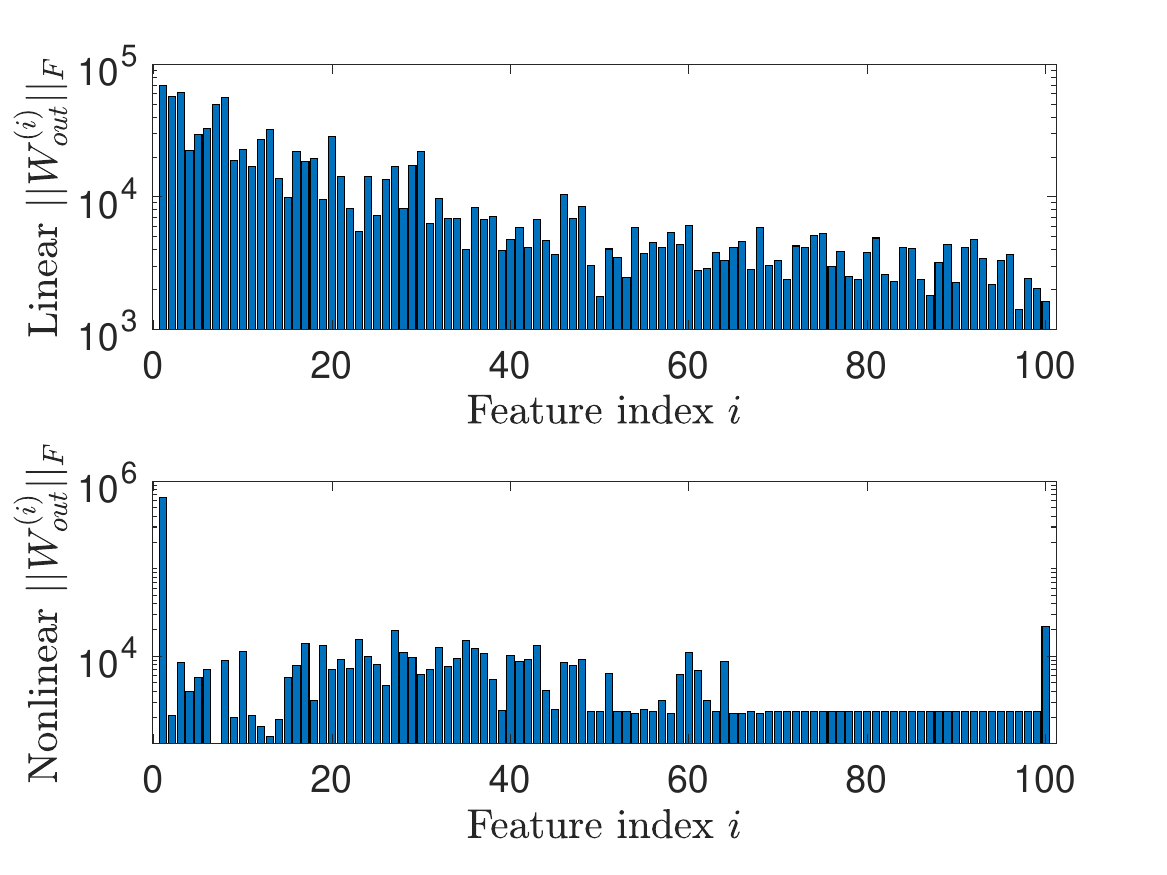}
\caption{Frobenius norm of the output map for different features of traffic-volume prediction} \label{Fig: NuminaOutput}
\end{figure}
\noindent where $\sigma=10$, $\rho=28$, and $\beta=8/3$ produces chaotic behavior. Table \ref{tb:hyparam} lists the hyperparameters used to train the ESN. The prediction via Feat-ESN and the regular ESN is depicted in Fig.~\ref{Fig: LorenzPrediction}. Fig.~\ref{Fig: LorenzError} provides a detailed error profile for different block-size $b$. The reservoir-size $n$ for the regular ESN is kept $n=N_fb$, i.e., same with the Feat-ESN, where the feature size $N_f$ for Lorenz system is 7. The normalized root mean square error (NRMSE) between the true sequence $\{\mf{x}(t_k): i=1,\ldots, l\}$ and the predicted sequence $\{\hat{\mf{x}}(t_k): i=1,\ldots, l\}$ is given by
\bql \footnotesize
\operatorname{NRMSE}(\mf{x},\hat{\mf{x}}) = \sqrt{\left(\sum\limits_{k=1}^l \norm{\mf{x}(t_k)-\hat{\mf{x}}(t_k)}^2\right)/\left(\sum\limits_{k=1}^l \norm{\mf{x}(t_k)}^2\right)},
\eql where $l$ is the prediction length. The prediction NRMSEs for different block-sizes over 50 independent Monte-Carlo trials are plotted in Fig.~\ref{Fig: LorenzError}. The performance of Feat-ESN algorithm remains consistent with different block-size $b$ while the performance of the regular ESN catches up with larger reservoir-size. This demonstrates the prediction capability of Feat-ESN with smaller reservoir-complexity. The contribution from each feature to the output is quantified by the Frobenius norm the corresponding output matrix $W_{out}^{(i)}$ for both the linear $\mf{r}$ and nonlinear $\psi(\mf{r})$ part and showed in Fig.~\ref{Fig: LorenzOutput}. The contribution from nonlinear terms is higher as expected.

\subsection{R\"ossler System}
Next, Feat-ESN is utilized to predict the time-series data generated by the R\"{o}ssler system described in \cite{Rossler1976}:
\bnl\label{Eq: Rossler}
\dot{x} &=& -y-x\\\nonumber
\dot{y} &=& x + ay \\\nonumber
\dot{z} &=& b + z(x-c),
\enl
with $a=0.5$, $b=2$, and $c=4$ to produce chaotic behavior. Similar to the Lorenz system example, the training data is corrupted by a measurement noise $\mf{v}(t_k) \sim \mc{N}(0, \sigma_v^2I_{3\times 3})$. Table \ref{tb:hyparam} lists the hyperparameters used to train the ESN. The prediction via Feat-ESN and regular ESN is depicted in Fig.~\ref{Fig: RosslerPrediction}. Fig.~\ref{Fig: RosslerError} plots the detailed error profile for different block-size $b$. The results are generated by 50 independent Monte-Carlo trials for training and testing the ESNs. The contributions from different features are also shown in Fig.~\ref{Fig: RosslerOutput}
\begin{table*}[t]
\vspace{5pt}
\footnotesize
\begin{center}
\caption{ESN hyperparameters}\label{tb:hyparam}
\begin{tabular}{lccc}
\hline
Hyperparameter & &Value &  \\
 & Lorenz system \eqref{Eq: Lorenz} & R\"{o}ssler system \eqref{Eq: Rossler} & Traffic Volume \\\hline
 Time step $\Delta t$ & $0.02$s & $0.1$s & $1$h\\
Block size $b$ & $5$ & $5$ & $10$\\
Feature size $N_f$ & $7$ & $7$ & $100$\\
Reservoir connection probability $p$ & $0.01$ & $0.01$ & $0.01$ \\
Training length $N$ & $5000$ & $1000$ & $1000$ \\ 
Nonlinear readout $\psi(\mf{r})$ & $\mf{r}^2$ & $\mf{r}^2$ & $\tanh(\mf{r})$\\
Leaking rate $\alpha$ & $0.3$ & $0.3$ & $0.7$\\
Regularization $\beta$ & $10^{-6}$ & $10^{-6}$ & $10^{-6}$\\\hline
\end{tabular}
\end{center}
\end{table*}
\subsection{Prediction of Traffic Volume on an Intersection of a Road Network}
Feat-ESN is now utilized for prediction of traffic volumes in road-intersections at different hours of the day. The network is trained on a dataset of traffic volumes obtained from Numina \cite{Numina} sensors at five different intersections on the University of Maryland campus. Fig.~\ref{Fig: TrafficSchematic}(a) represents the road network marked with sensor locations. Each sensor counts the number of pedestrians, bicycles, and vehicles at the respective intersections and store them in a server. The time series data of hourly vehicle traffic volume for two months is used. Fig.~\ref{Fig: TrafficSchematic}(b) represents the hourly vehicle traffic volume over a week with a clear daily pattern.

The traffic volume data originate from an infinite-dimensional spatio-temporal dynamical system evolving over a road network, and hence, the each sensor-recording provides a partial measurement. An ESN usually requires full state measurements in the training phase \cite{Goswami2021}, \cite{Lu2017}. A delay-embedding in the input layer \cite{Goswami2023} is used to account for the partial observation for training. The delay-embedding for Feat-ESN is demonstrated in Fig.~\ref{Fig: DelayFeatESN}. In this case, the features are chosen as the possible delay-combinations between $1$ to $m$, i.e., $N_f=m$. Both regular ESN and Feat-ESN is applied on this delay-embedded time series. Only a scalar measurement from sensor 1 is used in this paper with embedding dimension $m=100$.

The ESN is trained on 1000 hours of traffic volume data and tested for 70 hours, i.e., approximately three days. The training hyperparameters are listed in Table \ref{tb:hyparam}. Fig. \ref{Fig: NuminaPrediction} shows the traffic volume prediction by Feat-ESN and regular ESN. Fig.~\ref{Fig: NuminaError} shows the NRMSE and Pearson correlation coefficient between predicted and true traffic volumes with sensor data from intersection 1. The results are similar for the other four intersections and not included here. The Pearson correlation coefficient between true and predicted sequences ($\{x(i): i=1,\ldots, l\}$ and $\{\hat{x}(i): i=1,\ldots, l\}$ respectively) measures their normalized linear correlation. It is given by
\bql\footnotesize
r(x,\hat{x}) = \frac{\sum\limits_{k}\left(x(t_k)-\bar{x}\right)^T\left(\hat{x}(t_k)-\bar{\hat{x}}\right)}{\sqrt{\sum\limits_{i}\norm{x(t_k)-\bar{x}}^{2}} \sqrt{\sum\limits_{k}\norm{\hat{x}(t_k)-\bar{\hat{x}}}^{2}}},
\eql
where $\bar{x}$ and $\bar{\hat{x}}$ denotes the time-average values of $x(t_k)$ and $\hat{x}(t_k)$. Feat-ESN yields improved NRMSE and higher Pearson correlation coefficient with smaller block-size, i.e., with less number of reservoir nodes. The contribution from each feature, i.e., the number of delayed inputs for this case, is plotted in Fig.~\ref{Fig: NuminaOutput}. The contribution decreases with the increasing delay as expected.


\section{Conclusion}
This paper proposes a feature-based systematic approach to generate the reservoir for an echo-state network (ESN) that utilizes the power of smaller linear reservoirs fed with bite-sized input-features training it with nonlinear readout maps. The algorithm, called feature-based esn (Feat-ESN) uses parallel smaller linear neuronal reservoirs driven by different input combinations, called features in order to significantly reduce the computational complexity of the ESN while keeping the same predictive performance of a much larger reservoir. The proposed algorithm demostrates improved prediction performance with less reservoir nodes over the regular ESN for chaotic time-series. The method is then applied to a real data set of traffic patterns on the road network of the University of Maryland, College Park campus to predict the traffic volume at various intersections.

\section*{Acknowledgement}
The author thanks Dr. Derek A. Paley and the University of Maryland Department of Transportation for the Numina sensor data and Dr. Artur Wolek for preprocessing the data.
\bibliographystyle{IEEEtran}
\bibliography{bibl}

\end{document}